\documentclass[letterpaper]{article} 
\usepackage{aaai24}  
\usepackage{times}  
\usepackage{helvet}  
\usepackage{courier}  
\usepackage[hyphens]{url}  
\usepackage{graphicx} 
\usepackage{natbib}  
\usepackage{caption} 
\usepackage{algorithm}
\usepackage{algorithmic}
\usepackage{subfigure}
\usepackage{amsmath}
\usepackage{amssymb}
\usepackage{booktabs}
\usepackage[utf8]{inputenc} 
\usepackage[T1]{fontenc}    
\usepackage{multirow}
\usepackage{soul}
\usepackage{color}
\usepackage{xcolor}

\UseRawInputEncoding
\usepackage{newfloat}
\usepackage{listings}
\DeclareCaptionStyle{ruled}{labelfont=normalfont,labelsep=colon,strut=off} 
\floatstyle{ruled}
\newfloat{listing}{tb}{lst}{}
\floatname{listing}{Listing}
\title{HEAP: Unsupervised Object Discovery and Localization with \\Contrastive Grouping}
\author{
	Xin Zhang\textsuperscript{\rm 1}, 
	Jinheng Xie\textsuperscript{\rm 1}, 
	Yuan Yuan\textsuperscript{\rm 2},
	Michael Bi Mi\textsuperscript{\rm 2}, 
	Robby T. Tan\textsuperscript{\rm 1}
}
\affiliations{
    \textsuperscript{\rm 1}National University of Singapore \\
 	\textsuperscript{\rm 2}Huawei International Pte Ltd\\

    \{x.zhang, jinheng\}@u.nus.edu, yuanyuan10@huawei.com, michaelbimi@yahoo.com, robby.tan@nus.edu.sg
}

\begin{document}

\maketitle

\begin{abstract}
	Unsupervised object discovery and localization aims to detect or segment objects in an image without any supervision. 
	Recent efforts have demonstrated a notable potential to identify salient foreground objects by utilizing self-supervised transformer features.
	However, their scopes only build upon patch-level features within an image, neglecting region/image-level and cross-image relationships at a broader scale.
	Moreover, these methods cannot differentiate various semantics from multiple instances. 
	To address these problems, we introduce \textbf{H}ierarchical m\textbf{E}rging framework via contr\textbf{A}stive grou\textbf{P}ing (HEAP).
	Specifically, a novel lightweight head with cross-attention mechanism is designed to adaptively group intra-image patches into semantically coherent regions based on correlation among self-supervised features. 
	Further, to ensure the distinguishability among various regions, we introduce a region-level contrastive clustering loss to pull closer similar regions across images. 
	Also, an image-level contrastive loss is present to push foreground and background representations apart, with which foreground objects and background are accordingly discovered. 
	HEAP facilitates efficient hierarchical image decomposition, which contributes to more accurate object discovery while also enabling differentiation among objects of various classes. 
	Extensive experimental results on semantic segmentation retrieval, unsupervised object discovery, and saliency detection tasks demonstrate that HEAP achieves state-of-the-art performance.
\end{abstract}

\section{Introduction}

Unsupervised object discovery and localization aims to detect or segment salient objects in images without any human supervision~\cite{vo2020toward}. 
It is a fundamental problem since it provides a valuable source of weak supervision beneficial for various downstream tasks~\cite{zhang2023adaptive,jin2022unsupervised}, such as object detection~\cite{shin2022unsupervised,Xie_2021_ICCV} and semantic segmentation~\cite{xie2022c2am,li2023acseg,Xie_2022_CVPR}, etc.
Due to the absence of prior knowledge regarding visual characteristics or expected class membership, localizing possible objects within a scene presents significant challenges.

\begin{figure}[ht!]
	\centering
	\includegraphics[width=0.9\linewidth,trim={0 0, 0, 0},clip]{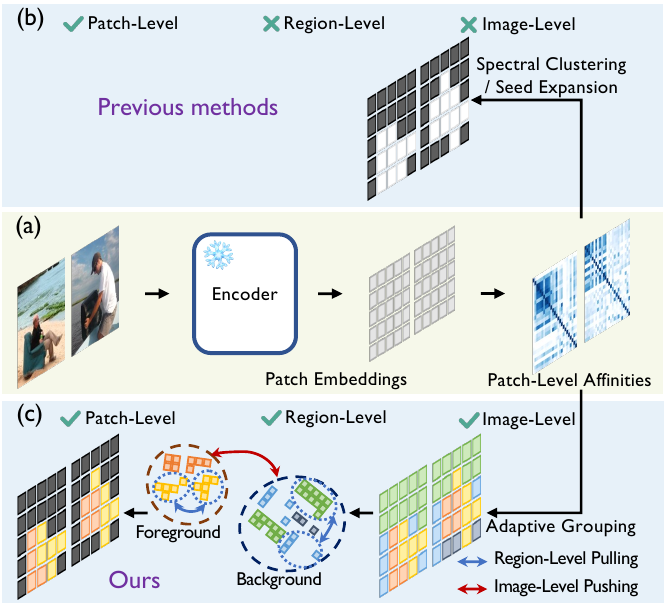}
	\caption{(a)-(b): Existing methods, such as FOUND~\cite{simeoni2023found} and TokenCut~\cite{wang2022self}, consider patch-level similarities for each image separately. (a)-(c): HEAP jointly explores hierarchical supervisions (i.e., patch/region/image-level) across images, enabling more acurate object discovery and discrimination among objects of various classes. Note that colors only represent different objects, not specific classes.}
	\label{fig.1}
\end{figure}

Early efforts~\cite{zitnick2014edge,uijlings2013selective,vo2019unsupervised,vo2021large} attempted to extract excessive candidate region proposals from the entire image set, which are computationally prohibitive to scale to large datasets.
More recent works~\cite{melas2022deep,wang2022self} turn to leveraging self-supervised models such as DINO~\cite{caron2021emerging}, which show great potential for object segmentation. 
\cite{melas2022deep,wang2022self} construct a similarity graph for each image separately and exploit spectral clustering to localize objects. 
\cite{shin2022unsupervised,wang2022freesolo,wang2023cut} further utilize the identified objects as pseudo masks to train a detection/segmentation model. 
\cite{simeoni2021localizing,simeoni2023found} first identify foreground/background seeds from the self-attention map and then expand to cover the whole object.
While yielding promising outcomes, existing methods treat the problem as binary patch-wise classification, ignoring semantic relationships between objects.
The investigated similarity is restricted within a single image and among individual patches, which might not adequately encapsulate the holistic representation of an entire object (Fig.~\ref{fig.1} (a)-(b)). 
Incorporating feature relationships at a broader scale (e.g., region/image-level) across images would be beneficial.

To this end, we propose a novel \textbf{H}ierarchical m\textbf{E}rging framework via contr\textbf{A}stive grou\textbf{P}ing (HEAP), to gradually establish patch-level, region-level, and image-level ties.
The core idea is that an image can be decomposed into semantically consistent regions (e.g., sky, land, and bird, etc), and objects are typically identified as salient foreground regions (Fig.~\ref{fig.1} (a)-(c)).
It has been studied that visual features extracted by recent self-supervised models highlight salient foreground objects while remaining semantically discriminative among various objects~\cite{caron2021emerging}. 
This motivates us to develop a lightweight head that takes in self-supervised patch features. 
The head then undertakes two learning tasks for object discovery: (1) grouping patches to form various semantic coherent regions; (2) discriminating foreground objects from background regions.

While the concept of parsing images through grouping has been explored in prior representation learning studies~\cite{wen2022self}, it often requires to learn a huge number of group prototypes that represent various concepts shared across the whole dataset. 
Such an approach has been proven to be less flexible for scalability.
Instead, we introduce a set of learnable group tokens, which adaptively interact with patch embeddings and become image-specific cluster centers for grouping patches. 
As such, we do not need to estimate the number of classes that a training set may contain, but the maximum number of regions that an image can be partitioned into.
This significantly alleviates the associated complexity.
It should also be pointed out that although GroupViT~\cite{xu2022groupvit} takes a similar idea to leverage group tokens in designing a segmentation model, it has to rely on text supervision for training the model from scratch.  
Conversely, HEAP is built upon a pre-trained encoder and can be efficiently trained without any external sources of supervision. 
Since the main purpose of this work is to design a universal head to manipulate any pre-trained features, a second-stage self-training that is commonly employed in some previous studies~\cite{shin2022unsupervised,wang2022freesolo,wang2023cut} is not considered.

To achieve accurate grouping and foreground-background disentanglement, we introduce three unsupervised losses, i.e., intra-image grouping, foreground-background disentanglement, and inter-image clustering losses.
The intra-image grouping loss enhances discrimination within each image by encouraging the grouping of similar patches. 
Consequently, objects with coherent semantics are grouped into different regions. 
This by nature helps distinguish multiple-class objects within a single image, which cannot be handled by previous methods. 
After grouping, patch embeddings with the same group token are merged to create region-level features. These region-level features are then aggregated to determine foreground objects from background regions.
Our intuition is that foreground objects across images share class similarities, and that background regions are more similar to backgrounds in other images than to foreground objects~\cite{xie2022c2am,simeoni2021localizing}.
%
To achieve this, we propose a cross-image foreground-background contrastive loss to differentiate foreground objects from background regions.
Also an additional inter-image clustering loss is present to promote consistent grouping across different images by encouraging the grouping of patches with similar semantics together across the dataset.

Following FOUND~\cite{simeoni2023found}, we extensively evaluate the proposed HEAP on semantic segmentation retrieval on VOC12~\cite{everingham2012pascal}, unsupervised saliency detection on DUT-OMRON~\cite{yang2013saliency}, DUTS-TE~\cite{wang2017learning}, and ECSSD~\cite{shi2015hierarchical}, and unsupervised object discovery on VOC07 \& VOC12~\cite{everingham2007pascal,everingham2012pascal}, and COCO20k~\cite{lin2014microsoft}. HEAP demonstrates the best performance while keeping the training at a low cost. Main contributions are:
\begin{itemize}
	\item We propose a hierarchical contrastive grouping framework, HEAP, for unsupervised object discovery and localization. HEAP explores representation learning at multiple levels which enables more accurate discovery and has the capacity to distinguish objects of multiple classes in a single image. 
	\item By optimizing the proposed head with the multi-level losses, this paper provides an efficient way to parse images hierarchically with no supervision. 
	\item Extensive experimental results on semantic segmentation retrieval, unsupervised object discovery and saliency detection tasks demonstrates that HEAP achieves state-of-the-art performance.
\end{itemize}

\section{Related Work}

\paragraph{Self-Supervised Methods}
Self-supervised methods have emerged as powerful approaches for leveraging unlabeled data to learn meaningful representations~\cite{he2020momentum,tian2020makes}. Recently, there has been a surge of research focused on vision transformer (ViT)-based self-supervised approaches~\cite{caron2021emerging,he2022masked,zhou2021ibot}. 
MoCo-v3~\cite{chen2021empirical} extends MoCo~\cite{he2020momentum} to incorporate ViTs, demonstrating effective self-supervised learning of ViTs on large-scale datasets.
DINO~\cite{caron2021emerging} employs instance discrimination in a teacher-student setup. The teacher network assigns pseudo-labels to the student network by comparing image patches, facilitating the learning of meaningful features. 
iBOT~\cite{zhou2021ibot} incorporates an online tokenizer for masked prediction, utilizing self-distillation on masked patch tokens with the teacher network as the tokenizer.
MAE~\cite{he2022masked} employs a high masking ratio on input images and an asymmetric encoder-decoder architecture for pixel-wise reconstruction.

\paragraph{Vision Transformer}
Vision Transformers (ViTs) have gained significant attention in recent years as a powerful alternative to convolutional neural networks (CNNs). 
The seminal work of~\cite{dosovitskiy2020vit} introduced the ViT architecture, demonstrating its efficacy in image classification by leveraging self-attention mechanisms. 
Since then, numerous studies have explored and extended the vision transformer framework for image classification~\cite{liu2021swin,touvron2021going}, object detection~\cite{li2022exploring,carion2020end,zhang2021vit,zhang2022dino,li2022mask} and segmentation~\cite{xie2021segformer,strudel2021segmenter,cheng2022masked}. 
GroupViT~\cite{xu2022groupvit} tackles semantic segmentation with text guidance. It breaks down images into meaningful regions, then employs contrastive learning between text and fine-grained region embeddings.
Our approach also uses a grouping mechanism, but unlike GroupViT, we don't rely on text supervision. We exclusively use contrastive learning with images, enabling the discovery of meaningful regions and objects in an unsupervised manner.

\paragraph{Unsupervised Object Discovery and Localization}
In the field of unsupervised object discovery and localization, objects are usually detected using binary masks~\cite{li2015weighted,vicente2011object,joulin2012multi} or bounding boxes~\cite{zhu2014saliency}.
Early methods used saliency or intra\&inter-image similarity for region proposals~\cite{zitnick2014edge, uijlings2013selective, wei2019unsupervised, vo2019unsupervised, vo2021large}, employing combinatorial optimization for object bounding box selection. 
However, their computational complexity hindered scalability for large datasets.
Recent works use pre-trained ViT models to extract features and construct a weighted graph with patches as nodes and pairwise similarities as edges.
LOST~\cite{simeoni2021localizing} incorporates a heuristic seed expansion strategy based on the graph. 
FOUND~\cite{simeoni2023found} takes a similar way as LOST but instead expands background patches.
Some works~\cite{shin2022unsupervised,melas2022deep,wang2022self} apply spectral clustering or normalized graph-cut to find the highly connected patches as objects. 
These methods ignore the relationships among images, leading to less optimal results. 
To tackle this, we introduce a hierarchical framework that groups patches into semantically consistent regions for object discovery. This approach identifies distinct foreground objects with varied semantics across regions.

\section{Proposed Method} 
\label{method}

\subsection{Architecture}
\begin{figure*}[ht!]
	\centering
	\includegraphics[width=0.9\linewidth,trim={0 0, 0, 0},clip]{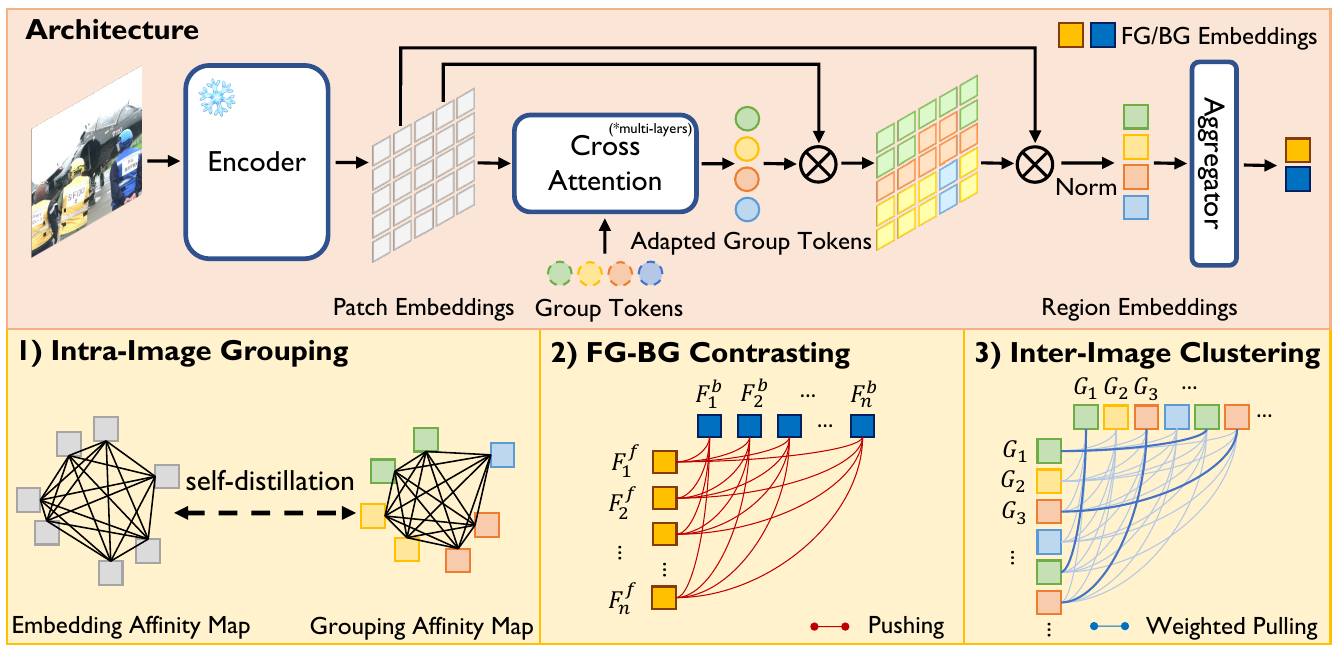}
	\caption{HEAP Overview. A pre-trained encoder (e.g., with DINO~\cite{caron2021emerging}) processes input images for patch embeddings. The cross-attention layers take in learnable group tokens and patch embeddings, and then adaptively aggregate image representations with each group token capturing distinct characteristics. Learned tokens cluster patches into regions based on embedding similarity. Patches belonging to the same regions are merged to form region-level embeddings, further aggregated for image-level foreground and background embeddings. HEAP is trained with three losses. Intra-image grouping: grouping patches based on similarities. Foreground-background contrasting: pushing foreground and background embeddings apart. Inter-image clustering: pulling similar regions closer, with the strength weighted by the similarity ranking. Note that $F^{f}_{i}$ represents the foreground embedding and $F^{b}_{i}$ represents the background embedding. $G_{i}$ can be $G^{f}_{i}$ or $G^{b}_{i}$, representing the region-level foreground or background embedding. We omit the process of obtaining $G^{f}_{i}$ and $G^{b}_{i}$ from region-level embeddings, which simplifies the illustration and does not affect understanding. }
	\label{fig:main}
\end{figure*}

As shown in Fig.~\ref{fig:main}, HEAP is built upon a frozen pre-trained encoder. It comprises a grouping block with multiple cross-attention layers, and a linear aggregator. The cross-attention layers take trainable group tokens and patch embeddings as inputs. The resulting adapted group tokens are used to assign individual patches to each group. Then region-level features are obtained by averaging patch embeddings of each group, which are then input into the linear aggregator for foreground-background disentanglement. 

Given a batch of $n$ images $\mathbf{I}_{1:n}=\{\mathbf{I}_{i}\}^{n}_{i=1}$, where $\mathbf{I}_{i}\in\mathbb{R}^{H\times W\times3}$, $H\times W$ denotes the spatial dimension. We first extract patch embeddings using a self-supervised pre-trained ViT. The image $\mathbf{I}_{i}$ is divided into $N$ non-overlapping image patches with patch resolutions of $K\times K$, i.e., $N=HW/K^{2}$. The $N$ patches are represented as $N$ patch tokens, along with an additional class token $\mathit{CLS}$. We extract patch embeddings $\mathbf{P}_{1:N}=\{\mathbf{P}_{i}\}^{N}_{i=1}$, in which $\mathbf{P}_{i}\in\mathbb{R}^{1\times D}$ from the last self-attention layer of the ViT by concatenating multiple attention heads, resulting in a feature dimention $D$ for each patch embedding. Note that, the $\mathit{CLS}$ token is not used. As such, each image $\mathbf{I}$ is represented as a collection of patch embeddings $\{\mathbf{P}_{i}\}_{i=1}^{N}$.

A set of $M$ learnable group tokens $\mathbf{g}_{1:M}=\{\mathbf{g}_{i}\}_{i=1}^{M}$ are initialized, in which $\mathbf{g}_{i}\in \mathbb{R}^{1\times D}$. The cross-attention is applied where group tokens serve as the \textsl{query}, and the concatenation of group tokens and patch embeddings as \textsl{key} and \textsl{value}. It involves interactions among group tokens and between group tokens and patch embeddings. The softmax-based attention mechanism inhibits one group token’s response to others and fosters competition among group tokens for information aggregation. Let $\mathbf{P}\in \mathbb{R}^{N\times D}$ denotes all the patch embedding of an image, $\mathbf{g}\in \mathbb{R}^{M\times D}$ denotes all the group tokens. The calculation of cross-attention follows the standard procedure of self-attention mechanism:
\begin{eqnarray}
	\label{eq:cross_attn1}
	\overline{\mathbf{g}}&=&{\rm softmax}(\frac{\mathbf{g}\mathbf{W_{q}}([\mathbf{g};\mathbf{P}]\mathbf{W_{k}})^T}{\sqrt{D}})([\mathbf{g};\mathbf{P}]\mathbf{W_{v}}), \\
	\label{eq:cross_attn2}
	\mathbf{g} &=& \mathbf{g} + \overline{\mathbf{g}}\mathbf{W}_{o},
\end{eqnarray}
where $\mathbf{W}_{q}$, $\mathbf{W}_{k}$, $\mathbf{W}_{v}$, $\mathbf{W}_{o}$ are the projection matrices of the cross-attention layer. By using the cross-attention, group tokens globally interact with other group tokens and patch embeddings. Consequently, the learned group tokens adaptively capture distinct semantic information. Then, each patch will be assigned to the  groups based on similarity in the embedding space. We compute the assignment matrix $\mathbf{Z}$ between the patch embeddings $\{\mathbf{P}_{i}\}$ and the group tokens $\{\mathbf{g}_{j}\}$ using \texttt{Gumbel-Softmax}:
\begin{equation}
	\mathbf{Z}_{i,j}=\frac{\exp(\mathbf{P}_{i}{\mathbf{g}_{j}}^T+\gamma_{i})}{\sum^{M}_{k=1}{\exp(\mathbf{P}_{i}\mathbf{g}_{k}^T+\gamma_{k})}},
\end{equation}
where $\gamma_{i}$,$\gamma_{k}$ are samples randomly drawn from the \texttt{Gumbel (0,1)} distribution. Then the assignment of patches to groups are determined by the maximum similarity:
\begin{equation}
	a_{i} = \mathop{\arg\max}_{j}(\mathbf{Z}_{i,j}).
\end{equation}
The region-level embeddings $\mathbf{G}_{1:M}=\{\mathbf{G}_{j}\}_{j=1}^{M}$ are then calculated by averaging all patch embeddings belonging to a specific group:
\begin{equation}
	\mathbf{G}_{j} = \frac{\sum_{i}^{N}\mathbb{I}(a_{i}==j)\mathbf{P}_{i}}{\sum_{i}^{N}\mathbb{I}(a_{i}==j)},
\end{equation}
where $\mathbb{I}$ is the indicator function. 
The aggregator $\phi(\cdot)$ is a linear layer that takes the region-level embedding $\mathbf{G}$ and outputs the probability of a group belonging to foreground. 
\subsection{Intra-Image Grouping}
HEAP takes an image as input and employs a region-splitting mechanism to partition it into multiple distinct regions, akin to superpixels. To achieve cohesive and perceptually meaningful grouping, we construct a fully connected undirected graph $\mathcal{G}$~=~($\mathcal{V}$, $\mathcal{E}$) where each $\mathcal{V}$ represents a patch embedding $\mathbf{P}_{i}$, and each patch is linked to all other patches by edges $\mathcal{E}$ representing the similarity scores based on the cosine similarity of the patch embeddings of the two patches.  As such, we build an affinity matrix $\mathbf{A}_{i,j}\in\mathbb{R}^{N\times N}$ of the $N$ patches. 
\begin{equation}
	\mathbf{A}_{ij} = \max(0, \cos (\mathbf{P}_{i}, \mathbf{P}_{j})),
\end{equation}
where $\cos$ calculates the cosine similarity.

\paragraph{Self-Distillation Loss} Intuitively, we expect similar patches to be grouped together. Inspired by~\cite{hamilton2022unsupervised,wang2022self}, we introduce a self-distillation loss that constrains the assigment probability of patches to groups $\mathbf{Z}$, with the affinity matrix of patch embeddings. Specifically, we define an affinity matrix $\delta_{ij}$ of assignment probability of each patch to groups as:
\begin{equation}
	\mathbf{\delta}_{ij} = \cos (\mathbf{Z}_{i\cdot}, \mathbf{Z}_{j\cdot}).
\end{equation}
Then, the goal of intra-image grouping is to distill the patch embedding affinity matrix $\mathbf{A}_{i,j}$ to the assignment probability affinity matrix $\delta_{ij}$. Taking the form of modularity in the field of community detection~\cite{newman2004finding}, the self-distillation loss is as follows:
\begin{equation}
	\mathcal{L}_{\rm intra} = -\frac{1}{2m}\sum\nolimits_{i,j}\left(\mathbf{A}_{ij}-\frac{k_{i}k_{j}}{2m}\right)\delta_{ij},
\end{equation}
where $2m=\sum_{i,j}\mathbf{A}_{ij}$, $k_{i}=\sum_{j}\mathbf{A}_{i,j}$. $m$ is used for normalization, and $k$ represent the degree of node $i$.
Note that, HEAP performs class-agnostic clustering, where group tokens lack specific semantics. Therefore, the self-distillation loss is computed image-wise. 
To prevent dominant groups and ensure balanced assignments, we add entropy regularization, maintaining reasonable average patch probabilities per group token. The final form of self-distillation loss is:
\begin{equation}
	\begin{split}
		\mathcal{L}_{\rm intra} = &-\frac{1}{2m}\sum\nolimits_{i,j}(\mathbf{A}_{ij}-\frac{k_{i}k_{j}}{2m})\delta_{ij} \\
		&+\lambda\left(\frac{1}{M}\sum\nolimits_{j}\left(\sum\nolimits_{i}{\mathbf{Z}_{ij}}\log\sum\nolimits_{i}{\mathbf{Z}_{ij}}\right)\right),
	\end{split}
\end{equation}
where $\lambda$ is the weight for the entropy regularization.

\subsection{Foreground-Background Disentanglement}\label{FBD}
Foreground objects are detected under the assumption that foreground and background regions have distinct representations. This leads to foreground features being more similar to other foreground representations across images, rather than to all background representations.
Given the region-level embeddings $\mathbf{G}$, the linear aggregator produces the probability of each region belonging to the foreground: $\mathbf{H}=\sigma(\phi(\mathbf{G}))$, where $\sigma$ is the sigmoid function and $\mathbf{H}\in\mathbb{R}^{M\times1}$. Based on $\mathbf{H}$, we can compute the foreground and the background representations, $\mathbf{F}^{f}\in\mathbb{R}^{1\times D}$ and $\mathbf{F}^{b}\in\mathbb{R}^{1\times D}$ as:
\begin{equation}
	\mathbf{F}^{f} = {\mathbf{H}^T}\otimes{\mathbf{G}}, 
	\mathbf{F}^{b} = (1-{\mathbf{H}^T})\otimes{\mathbf{G}}, 
\end{equation}
where $\otimes$ represents matrix multiplication. For a given batch of $n$ images, the aggregator produces corresponding $n$ foreground and $n$ background representations, i.e., $\{\mathbf{F}_{i}^{f}\}_{i=1}^{n}$ and $\{\mathbf{F}_{i}^{g}\}_{i=1}^{n}$. The objective is to push the two kinds of representations apart. To achieve this, we introduce a negative contrastive loss defined as follows:
\begin{equation}
	\mathcal{L}_{\rm neg} = -\frac{1}{n^2}\sum\nolimits_{i,j}\log(1-\cos(\mathbf{F}_{i}^{f}, \mathbf{F}_{j}^{b})),
\end{equation}
where $\cos(\mathbf{F}_{i}^{f}, \mathbf{F}_{j}^{b})$ computes the cosine similarity. The intention is to maximize the dissimilarity between all pairs of foreground and background representations in the batch.
\subsection{Inter-Image Clustering}
Relying solely on intra-image data may not ensure precise patch grouping. We introduce cross-image information, capturing contextual object relationships and appearances across images. This enriches understanding of object boundaries and enhances grouping accuracy. We introduce a cross-image similarity measure to quantify region correlations.

We aim to separately pull together similar foreground and background region representations. Specifically, for a given batch of $n$ images, the grouping block produces $n\times M$ image regions. For each of them, the aggregator outputs the probabilities for them to be the foreground, resulting in $\mathbf{H}\in\mathbb{R}^{nM \times 1}$. We have the region-level foreground and background representations expressed as:
\begin{equation}
	\mathbf{H}_{1:nM}=\sigma(\phi(\mathbf{G}_{1:nM})),
\end{equation}
where $\mathbf{G}_{1:nM}$ are the region-level embeddings of $n$ images, each image with $M$ regions. $\mathbf{H}_{1:nM}$ are the corresponding probabilities to be foreground. Then, we can have region-level foreground and background representations written as:
\begin{equation}
	\mathbf{G}^{f}_{i} = \mathbf{H}_{i}\odot\mathbf{G}_{i},
	\mathbf{G}^{b}_{i} = (1-\mathbf{H}_{i})\odot\mathbf{G}_{i},
\end{equation}
where $\odot$ represents element-wise multiplication and $i\in\{1,...,nM\}$. Subsequently, we introduce a ranking-based inter-image clustering loss that draws closer the similar region-level foreground/background representations according to the pairwise similarities. For each pair of region-level foreground/background representations, the similarity is defined as: $s_{i,j}^{f}=\max(0,\cos(\mathbf{G}_{i}^{f},\mathbf{G}_{j}^{f}))$, $s_{i,j}^{b}=\max(0,\cos(\mathbf{G}_{i}^{b},\mathbf{G}_{j}^{b}))$, where $i,j \in \{1, ..., nM\}$. The weight $w_{ij}$ is determined by the ranking on all possible pairs, i.e., $j\in \{1, ..., nM\}\backslash\{i\}$, and ranges from 0 to 1:
$
w_{i,j}^{f} = \exp(-\alpha\cdot\mathrm{rank}(s_{i,j}^{f})),
w_{i,j}^{b} = \exp(-\alpha\cdot\mathrm{rank}(s_{i,j}^{b})),
$
where $\alpha$ is a hyper-parameter controlling the weights.
The inter-image clustering loss is defined as: $\mathcal{L}_{\rm inter}=\mathcal{L}_{\rm inter\_fg} + \mathcal{L}_{\rm inter\_bg}$, where:
\begin{eqnarray}
	\nonumber
	\mathcal{L}_{\rm inter\_fg} &=& -\frac{1}{nM(nM-1)}\sum\nolimits_{i,j, i\neq j}(w_{i,j}^{f}\cdot\log(s^{f}_{i,j})),\\ \nonumber
	\mathcal{L}_{\rm inter\_bg} &=& -\frac{1}{nM(nM-1)}\sum\nolimits_{i,j, i\neq j}(w_{i,j}^{b}\cdot\log(s^{b}_{i,j})). \nonumber
\end{eqnarray}

The overall objective $\mathcal{L}$ is formulated as the summation of the three proposed losses:
\begin{equation}
	\mathcal{L} = \mathcal{L}_{\rm intra} + \mathcal{L}_{\rm neg} + \mathcal{L}_{\rm inter}.
\end{equation}

\section{Experiments}
\label{exp}

Implementation details are provided in the supplementary material. 
Dense CRF (dCRF)~\cite{krahenbuhl2011efficient} is utilized to refine the boundary of objects. 

\paragraph{Semantic Segmentation Retrieval}
\label{SSR}
We evaluate HEAP on unsupervised semantic segmentation retrieval on VOC12~\cite{everingham2012pascal}. We follow the common protocol in MaskContrast~\cite{van2021unsupervised} and compare with FreeSOLO~\cite{wang2022freesolo}, C2AM~\cite{xie2022c2am} TokenCut~\cite{wang2022self}, SelfMask~\cite{shin2022unsupervised} and FOUND~\cite{simeoni2023found}. We consider both single-object retrieval and multiple-objects retrieval setups. For single-object retrieval, all foreground objects are merged as one object while for multiple-objects retrieval, each foreground object is treated as a single object separately. Note that we also discard the foreground objects smaller than 1\% of an input image size. 

We build the feature bank on the train split and find the nearest neighbors of each object in the val set by retrieving the feature bank and assign them the corresponding ground-truth labels. We obtain the object-wise binary mask, followed by aggregating patch-level features within the masked region to generate object-level features. Mean Intersection-overUnion (mIoU) is used for evaluation.

\begin{figure*}[tbp] 
	\centering  
	\subfigure{
		\includegraphics[width=1.15in]{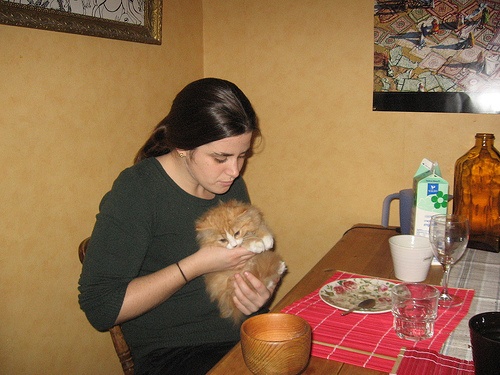}
	}
	\subfigure{
		\includegraphics[width=1.15in]{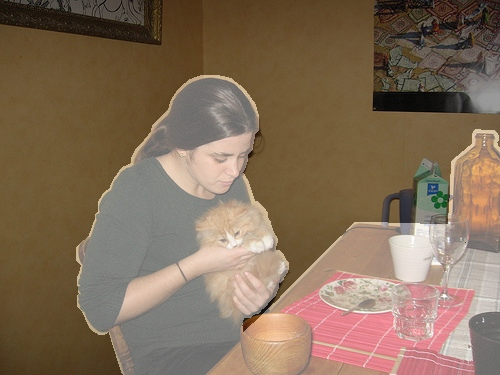}
	}
	\subfigure{
		\includegraphics[width=1.15in]{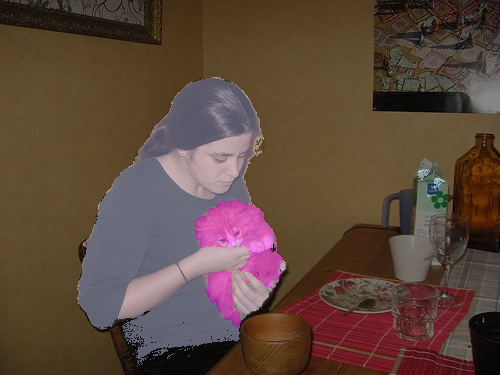}
	}
	\subfigure{
		\includegraphics[width=1.15in]{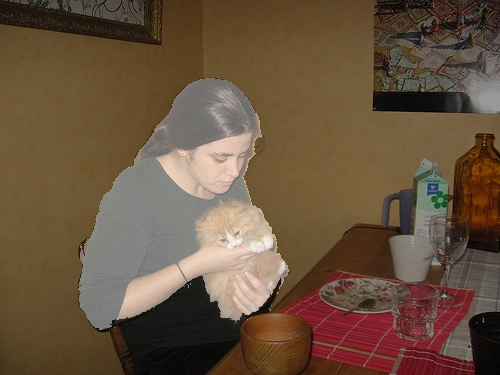}
	}
	\subfigure{
		\includegraphics[width=1.15in]{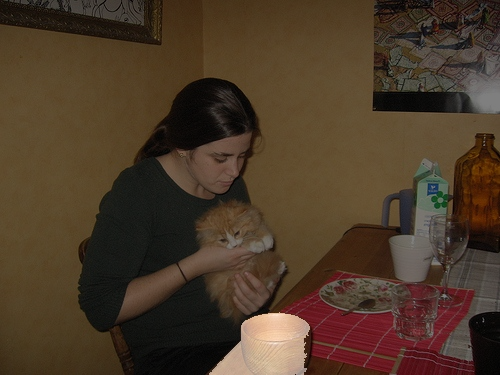}
	}
	\quad    
	\subfigure{
		\includegraphics[width=1.15in]{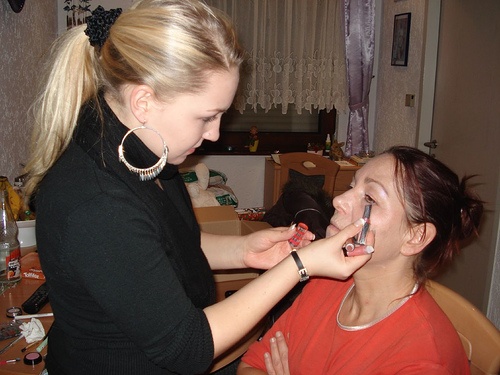}
	}
	\subfigure{
		\includegraphics[width=1.15in]{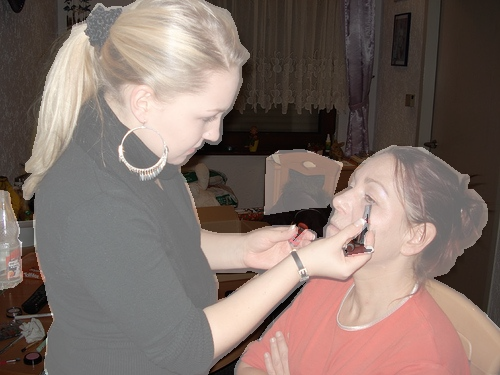}
		\label{label_for_cross_ref_1}
	}
	\subfigure{
		\includegraphics[width=1.15in]{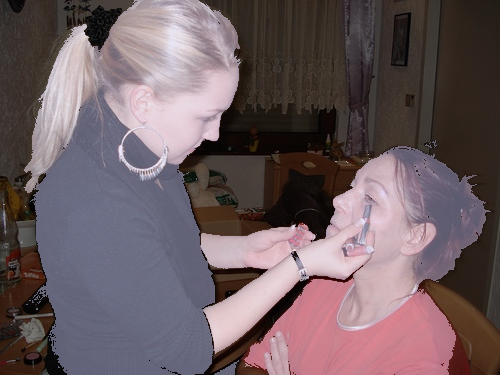}
		\label{label_for_cross_ref_1}
	}
	\subfigure{
		\includegraphics[width=1.15in]{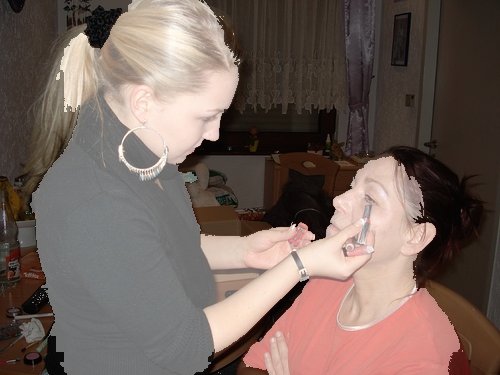}
	}
	\subfigure{
		\includegraphics[width=1.15in]{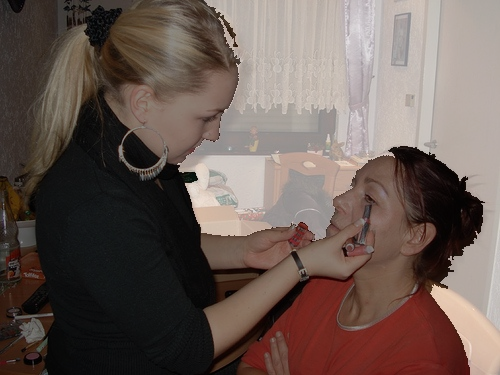}
	}
	\quad    
	\quad   
	\setcounter{subfigure}{0}
	\subfigure[Raw image]{
		\includegraphics[width=1.15in]{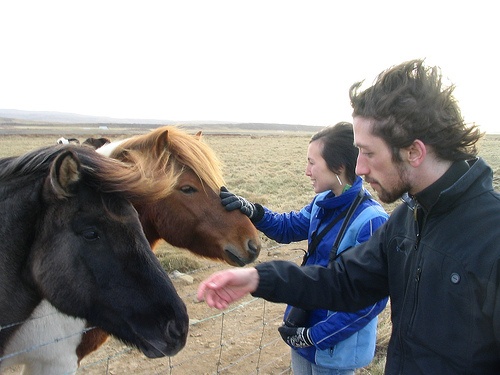}
	}
	\subfigure[Ground truth]{
		\includegraphics[width=1.15in]{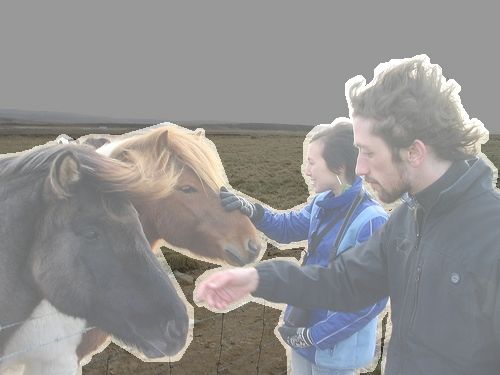}
	}
	\subfigure[HEAP (Ours)]{
		\includegraphics[width=1.15in]{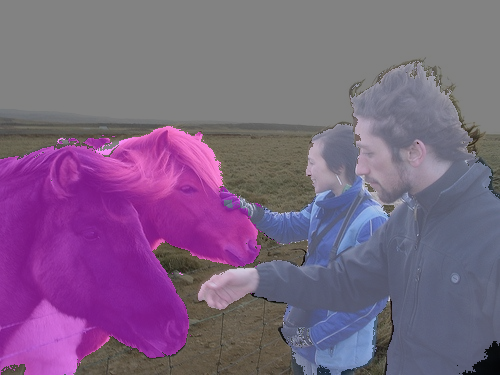}
	}
	\subfigure[FOUND]{
		\includegraphics[width=1.15in]{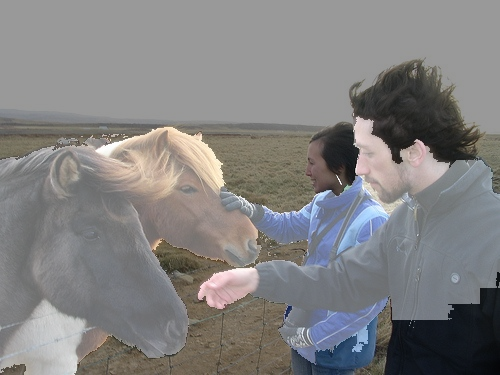}
	}
	\subfigure[TokenCut]{
		\includegraphics[width=1.15in]{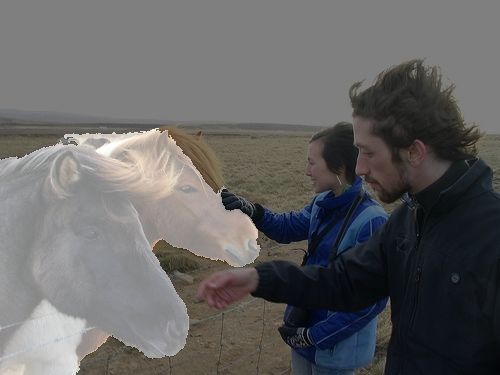}
	}
	\caption{Qualitative results of unsupervised object discovery and localization obtained by HEAP, FOUND~\cite{simeoni2023found}, and TokenCut~\cite{wang2022self} on VOC12. Our method could distinguish objects of multiple classes in a single image while recent works not. Note that colors only represent different objects in an image, not corresponding to specific classes. }
	\label{fig.2}
\end{figure*}

Table~\ref{tab:retrieval} summarizes the results. We borrow most results of existing methods from \cite{simeoni2023found}. Both cases where 7 (bus, airplane, car, person, cat, cow, and bottle) classes or all 21 classes of VOC12 are considered following \cite{van2021unsupervised}. It can be observed that HEAP consistently outperforms previous methods by a large margin on both single-object and multiple-objects retrieval tasks. Notably, HEAP surpasses FOUND by 3.3\% for multiple-objects retrieval when all the 21 classes are considered. This indicates the superiority of HEAP to discover objects of multiple categories in a single image. 

We also show and compare visualizations of the predicted saliency masks by HEAP, FOUND and TokenCut in Fig.~\ref{fig.2}. Note that post-processings (bilateral solver or dense CRF) are applied. It can be observed that TokenCut is prone to focus on a wrong region in an image (the 1nd and 2rd rows) and miss some objects (the 3th row). FOUND tends to undersegment small ratios of objects (the 2rd and 3th rows). However, HEAP yields more accurate masks. Moreover, objects of different classes are also successfully distinguished. 

\begin{table}[ht!]
	\centering
	\fontsize{9}{10}\selectfont
	{
		\begin{tabular}{lcc} 
			\toprule
			\multicolumn{1}{c}{} & \multicolumn{2}{c}{mIoU} \\ \cmidrule{2-3}
			Method & 7cls & 21cls \\
			\midrule
			MaskContrast (unsup. sal.)  & 53.4 & 43.3 \\
			\midrule
			\bf \textsl{Single saliency mask} \\
			FreeSOLO & 19.7 & 17.0 \\
			FreeSOLO (largest inst.) & 20.6 & 20.6 \\
			C2AM (ResNet50) & 45.4 & 35.4 \\
			TokenCut (ViT-S/8) & 46.7 & 37.6 \\
			TokenCut (ViT-S/16) & 49.7 & 39.9 \\
			SelfMask & 56.6 & 40.7 \\
			FOUND (ViT-S/8) & 56.1 & 42.9 \\
			\bf HEAP (ours) & \bf 59.1 & \underline{45.6} \\
			\midrule
			\bf \textsl{Multiple saliency masks} \\
			FreeSOLO & 23.9 & 25.7 \\
			SelfMask & 56.2 & 40.8 \\
			FOUND (ViT-S/8) & 58.0 & 42.7 \\
			\bf HEAP  (ours) & \bf 59.1 & \bf 46.0 \\
			\bottomrule
		\end{tabular}
	}
	\caption{Semantic segmentation retrieval on VOC12. Both single-object (the 2nd block) and multiple-object retrievals (the 3rd block) are considered.  MaskContrast uses its own trained feature extractor. Bold highlights the best results. Underlined represents the second best.}
	\label{tab:retrieval}
\end{table}

\begin{table}[h!]
	\centering
	\fontsize{9}{10}\selectfont
	{
		\begin{tabular}{lccc}
			\toprule
			Method & VOC07 & VOC12 & COCO20K \\
			\midrule
			LOD & 53.6 &  55.1 & 48.5 \\
			DINO-seg (ViT-S/16) & 45.8 & 46.2 & 42.0 \\
			LOST (ViT-S/8)  & 55.5 & 57.0 & 49.5 \\
			LOST (ViT-S/16) & 61.9 & 64.0 & 50.7 \\
			DSS (ViT-S/16) & 62.7 & 66.4 & 52.2 \\
			TokenCut (ViT-S/8) & 67.3 & 71.6 &  60.7 \\
			TokenCut (ViT-S/16) & 68.8 & 72.1 & 58.8 \\
			FreeSolo & 44.0 & 49.7 & 35.2 \\ 
			LOST* (ViT-S/16) & 65.7 & 70.4 & 57.5 \\
			TokenCut* (ViT-S/16) & 71.4 & 75.3 & 62.6 \\
			SelfMask & 72.3 & 75.3 & 62.7 \\
			DINOSAUR & --- & 70.4 & \bf 67.2 \\
			FOUND (ViT-S/8) &  \underline{72.5} & \underline{76.1} & 62.9 \\ 
			\textbf{HEAP (ours)} (ViT-S/8) & \bf 73.2 & \bf 77.1 & \underline{63.4} \\
			\bottomrule
		\end{tabular}
	}
	\caption{We compare HEAP to state-of-the-art object discovery methods on VOC07, VOC12, and COCO20K using CorLoc metric. * refers to a class-agnostic detector trained with unsupervised ``pseudo-boxes'' labels in the second stage. `ViT' represents different ViT architectures pre-trained with DINO. Bold highlights the best results, Underlined represents the second best. We provide the full table in supplementary materials.}
	\label{tab:obj_discovery}
\end{table}

\paragraph{Unsupervised Object Discovery}
We follow the common practice of~\cite{simeoni2023found,wang2022self} and evaluate HEAP on VOC07~\cite{everingham2007pascal}, VOC12~\cite{everingham2012pascal}, and COCO20K~\cite{lin2014microsoft,vo2020toward}. \textsl{CorLoc} metric is used to report results, which measures the proportion of images on which at least one object was correctly localized. Calculation of \textsl{CorLoc}  is detailed in supplementary materials. 

\begin{table*}[htbp]
	\centering
	\fontsize{9}{10}\selectfont
	{
		\begin{tabular}{lccccccccc}
			\toprule
			&  \multicolumn{3}{c}{DUT-OMRON} & \multicolumn{3}{c}{DUTS-TE} & \multicolumn{3}{c}{ECSSD} \\
			\cmidrule(rl){2-4}\cmidrule(rl){5-7}\cmidrule(rl){8-10}
			Method & Acc & IoU & max $F_\beta$ & Acc & IoU & max $F_\beta$ & Acc & IoU & max $F_\beta$ \\
			\midrule
			DSS & --- & .567 & --- & --- & .514 & --- & --- & .733 & --- \\
			LOST (ViT-S/16) & .797 & .410 & .473 & .871 & .518 & .611 & .895 & .654 & .758 \\
			SelfMask & .901 & \underline{.582} & --- & .923 & .626 & --- & .944 & .781 & --- \\
			TokenCut (ViT-S/16) & .880 & .533 & .600 & .903 & .576 & .672 & .918 & .712 & .803 \\
			FOUND (ViT-S/8) & \underline{.912} & .578 & \underline{.663} & \underline{.938} & \bf .645 & \underline{.715} & \bf .949  & \underline{.807} & \bf .955 \\
			\textbf{HEAP (ours)} (ViT-S/8) & \bf .920 & \bf .596 & \bf .690 & \bf .940 & \underline{.644} & \bf .757 & \underline{.945} & \bf .811 & \underline{.930} \\
			\midrule
			LOST (ViT-S/16) + BS &.818 &.489 &.578 &.887 &.572 &.697 &.916 &.723 &.837 \\
			SelfMask + BS & .919 & \bf .655 & --- & .933 & .660 & --- & \underline{.955} & \underline{.818} & --- \\
			TokenCut (ViT-S/16) + BS & .897 & .618 & .697 &.914 &.624 & .755 &.934 &.772 &.874 \\
			FOUND (ViT-S/8) + BS & \underline{.922} & .613 &  \underline{.708} & \underline{.942} & \underline{.663} & \underline{.763} & .951 & .813 & \underline{.935} \\
			\textbf{HEAP (ours)} (ViT-S/8) + dCRF & \bf .929 & \underline{.646} & \bf .724 & \bf .949 & \bf .687 & \bf .777 & \bf .962 & \bf .823 & \bf .948 \\
			\bottomrule
		\end{tabular}
	}
	\caption{We compare HEAP to state-of-the-art  methods on unsupervised saliency detection on DUT-OMRON, DUTS-TE, ECSSD. `+BS' refers to applying the post-processing bilateral solver on the generated masks; `+dCRF' refers to applying the post-processing dense CRF on the generated masks. Bold highlights the best results, underlined represents the second best.		
	}
	\label{tab:saliency-detection}
\end{table*}

Table~\ref{tab:obj_discovery} summarizes the quanlitative results. The results of most existing methods are from \cite{simeoni2023found}. We list comparisons with more recent self-supervised representation-based methods, e.g., LOST~\cite{simeoni2021localizing}, TokenCut, SelfMask, and FOUND, etc. Full table including more baselines is provided in supplementary materials. We observe that HEAP surpasses all existing methods on VOC07 and VOC12. Although DINOSAUR obtains higher performance on COCO20K, the training is much heavier, requiring 8 times larger GPU memory and hundreds times more training steps. Conversely, we aim to introduce an efficient and lightweight head that can also generalize to other pre-trained backbones effortlessly.

\paragraph{Unsupervised Saliency Detection}
We evaluate HEAP on three commmonly used datasets for unsupervised saliency detection: ECSSD~\cite{shi2015hierarchical}, DUTS-TE~\cite{wang2017learning}, and DUT-OMRON~\cite{yang2013saliency}. Following~\cite{simeoni2023found}, intersection-over-union (\textsl{IoU}), pixel accuracy (\textsl{Acc}) and maximal $F_{\beta}$ score (max $F_{\beta}$) with $\beta^2=0.3$ are used as metrics. Calculations are detailed in supplementary materials. 
Similar to \cite{simeoni2023found}, we consider the mask used for evaluation is with all identified objects. dCRF also processes all connected components. 

\begin{table}[ht]
	\centering
	\fontsize{9}{10}\selectfont
	\begin{tabular}{lccc|lccc}
		\toprule
		Num. & Acc & IoU & max $F_\beta$ & $\alpha$ & Acc & IoU & max $F_\beta$ \\
		\midrule
		3 & .913 & .604 & .736 & 0.1 & \bf .949 & \bf .687 & \bf .777 \\
		8 & \bf .949 & \bf .687 & \bf .777 & 0.25 & .925 & .659 & .756 \\
		15 & .933 & .654 & .758 & 0.5 & .931 & .656 & .762 \\
		\bottomrule
	\end{tabular}
	\caption{Effect of the number of group tokens $M$ and inter-image contrastive clustering loss weight $\alpha$. Experiments are evaluated on unsupervised saliency detection on DUTS-TE.}
	\label{M_alpha}	
\end{table}

\begin{table}[hbp]
	\centering
	\fontsize{9}{10}\selectfont
	{
		\begin{tabular}{ccccccc}
			\toprule
			$\mathcal{L}_{\rm intra}^*$ & ER & $\mathcal{L}_{\rm neg}$ & $\mathcal{L}_{\rm inter}$ & Acc & IoU & max $F_\beta$ \\
			\midrule
			\checkmark & & \checkmark & & .930 & .642 & .700 \\
			\checkmark & \checkmark & \checkmark & & .937 & .662 & .731 \\
			\checkmark & \checkmark & \checkmark & \checkmark & \bf .949 & \bf .687 & \bf .777 \\
			\bottomrule
	\end{tabular}}
	\caption{The effect of each loss. Experiments are evaluated on unsupervised saliency detection on DUTS-TE. Note $\mathcal{L}_{\rm intra}^*$ refers to the $\mathcal{L}_{\rm intra}$ without the entropy regularization, corresponding to Eq. 8. ER refers to the entropy regularization, corresponding to the second item in Eq. 9. }
	\label{loss}	
\end{table}

The results are shown in Table \ref{tab:saliency-detection}. Full table including more baselines is provided in supplementary materials. Compared with LOST, SelfMask, TokenCut, and FOUND, HEAP achieves better or comparable performance on all three datasets, indicating that HEAP effectively discovers objects through grouping and foreground-background representations contrasting. Similar to those methods, our approach does not involve training segmentation decoders. The patch-level predictions result in coarse segmentation. We improve this by using dCRF to refine object boundaries and reduce false positives, thereby improving performance.

\subsection{Ablation Studies}
All ablation experiments are conducted on unsupervised saliency detection on DUTS-TE.
\paragraph{Effect of Each Loss}
\label{EL}
We verify the effectiveness of each proposed loss. We start by the simplest version, which only implements intra-image grouping loss and foreground-background contrastive loss. The other losses are added one by one. Results are shown in Table~\ref{loss}. We can observe the performance gain brought by each loss. 

\paragraph{Effect of  Number of Group Tokens} $M$ determines the maximum number of regions (i.e., clusters) that an image can be divided into. The actual number of regions is adaptively determined for each image by an \texttt{argmax} operation. Theoretically, a small $M$ results in coarse segmentation; while a too large $M$ may lead to overclustering and segment objects into finer parts. As shown in the left part of Table~\ref{M_alpha}, HEAP maintains good performance under a range of $M$ and achieves best when $M=8$. 

\paragraph{Effect of $\alpha$} $\alpha$ controls the impact of neighboring samples in inter-image region-level clustering. A higher $\alpha$ prioritizes nearest neighbors, while a smaller $\alpha$ considers a broader range of neighbors. From the right part of Table~\ref{M_alpha}, we can observe that $\alpha=0.1$ results in best performance. Since region-level embedding space across images are much more diverse than that within a single image, it is natural to rely on more neighbourhood regions for conducting clustering. 
\section{Conclusion}

We propose HEAP, a novel approach that addresses unsupervised object discovery and localization. 
HEAP achieves this by hierarchically merging self-supervised local patterns to construct global representations of foreground objects and backgrounds.  
Through adaptive grouping and contrastive clustering, discriminative regions are formed, with which objects are identified by cross-image foreground-background contrasting.
HEAP's hierarchical property benefits discovering more complete objects and distinguishing objects belonging to multiple categories.
As a lightweight and general head, HEAP achieves state-of-the-art results on semantic segmentation retrieval, unsupervised object discovery and unsupervised saliency detection. 

\bibliography{aaai24}

\end{document}